\documentclass[runningheads]{llncs}
\usepackage{graphicx}

\usepackage{amsmath}
\usepackage{amssymb}
\usepackage{mathtools, stmaryrd}
\usepackage{xcolor}
\usepackage{multirow}

\usepackage{wrapfig,lipsum,booktabs}

\usepackage[capitalize]{cleveref}
\crefname{section}{Sec.}{Secs.}
\Crefname{section}{Section}{Sections}
\Crefname{table}{Table}{Tables}
\crefname{table}{Tab.}{Tabs.}

\newcommand{\eg}{\textit{e.g.~}}

\DeclarePairedDelimiterX{\Iintv}[1]{\llbracket}{\rrbracket}{\iintvargs{#1}}

\newcommand{\V}[1]{{\boldsymbol{#1}}}
\newcommand{\Vh}[1]{{\hat{\boldsymbol{#1}}}}

\begin{document}
\NewDocumentCommand{\iintvargs}{>{\SplitArgument{1}{,}}m}
{\iintvargsaux#1} %
\NewDocumentCommand{\iintvargsaux}{mm} {#1\mkern1.5mu..\mkern1.5mu#2}
\title{Memory transformers for full context and high-resolution 3D Medical Segmentation}
\titlerunning{ }
\author{Loic Themyr\inst{1,2}\orcidID{0000-0003-1396-2383} \and
Clément Rambour \inst{1}\orcidID{0000-0002-9899-3201} \and
Nicolas Thome\inst{1}\orcidID{0000-0003-4871-3045} \and
Toby Collins\inst{2}\orcidID{0000-0002-9441-8306} \and
Alexandre Hostettler\inst{2}\orcidID{0000-0001-8269-6766}}
\authorrunning{L. Themyr et al.}

\institute{Conservatoire National des Arts et Métiers, Paris 75014, France \and
IRCAD, Strasbourg 67000, France \\
\email{loic.themyr@lecnam.net}}

\maketitle              


\begin{abstract}

Transformer models achieve state-of-the-art results for image segmentation. However, achieving long-range attention, necessary to capture global context, with high-resolution 3D images is a fundamental challenge. This paper introduces the Full resolutIoN mEmory (FINE) transformer to overcome this issue. The core idea behind FINE is to learn memory tokens to indirectly model full range interactions while scaling well in both memory and computational costs. FINE introduces memory tokens at two levels: the first one allows full interaction between voxels within local image regions (patches), the second one allows full interactions between all regions of the 3D volume. Combined, they allow full attention over high resolution images, \eg 512 x 512 x 256 voxels and above.
Experiments on the BCV image segmentation dataset shows better performances than state-of-the-art CNN and transformer baselines, highlighting the superiority of our full attention mechanism compared to recent transformer baselines, \eg CoTr, and nnFormer.
\end{abstract}
\section{Introduction}

Convolutional encoder-decoder models have achieved remarkable performance for medical image segmentation \cite{bakas2018identifying,KAMNITSAS201761}. U-Net \cite{ronneberger2015unet} and other U-shaped architectures remain popular and competitive baselines. However, the receptive fields of these CNNs are small, both in theory and in practice \cite{RF-NeurIPS16}, preventing them from exploiting global context information.  

Transformers witnessed huge successes for natural language processing {\cite{AIAYN,devlin2018bert}} and recently in vision for image classification \cite{dosovitskiy2020vit}. One key challenge in 3D semantic segmentation is their scalability
, since attention's complexity is quadratic with respect to the number of inputs.

Efficient attention mechanisms have been proposed, including sparse or low-rank attention matrices \cite{qiu2020blockwise,wang2020linformer}, kernel-based methods \cite{peng2020random,katharopoulos2020transformers}, window \cite{liu2021Swin,fan2021multiscale}, and memory transformers \cite{rae2019compressive,lee2019set}. 
Multi-resolution transformers~\cite{liu2021Swin,wang2021pvtv2,wang2021pyramid} apply attention in a hierarchical manner by chaining multiple window transformers. Attention at the highest resolution level is thus limited to local image sub-windows. The receptive field is gradually increased through pooling operations. 
Multi-resolution transformers have recently shown impressive performances for various 2D medical image segmentation tasks such as multi-organ \cite{karimi_2021,Valanarasu_2021,cao2021swinunet}, histopathological \cite{Li_2021}, skin \cite{wang_2021}
, or brain \cite{Reynaud_2021} segmentation. 
\begin{figure}[h!]
    \centering
    \includegraphics[width=0.9\linewidth]{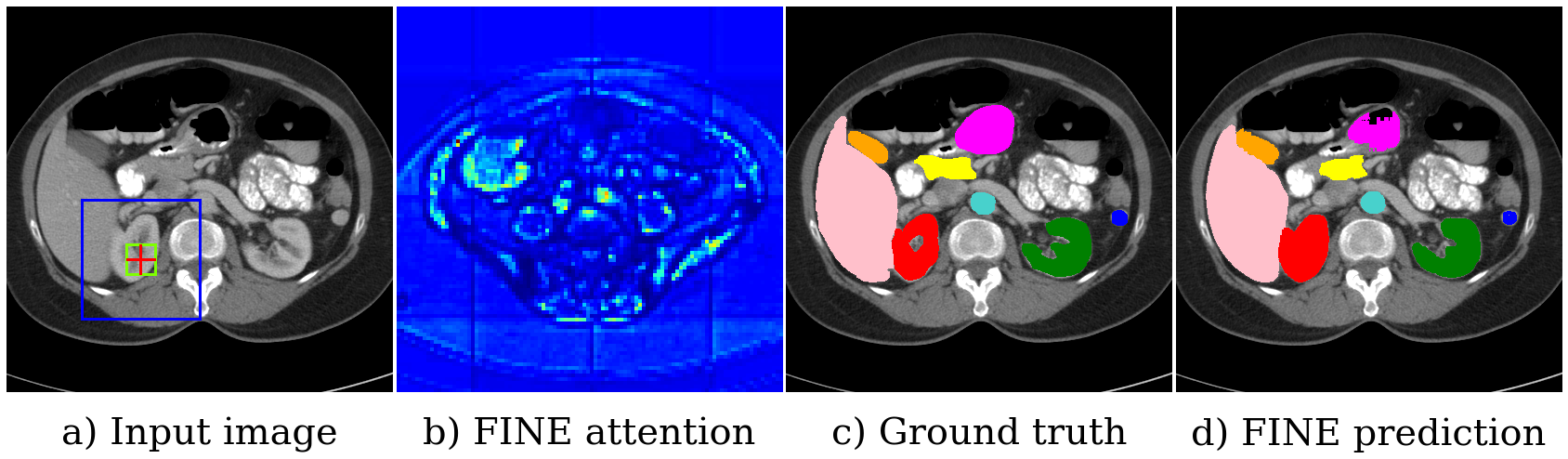}
    \caption{Proposed full resolution memory transformer (FINE). To segment the kidney voxel in a) (red cross), FINE combines high-resolution and full contextual information, as shown in the attention map in b). This is in contrast to nnFormer \cite{zhou2021nnformer} (resp. CoTr \cite{cotr}), which receptive field is limited to the green (resp. blue) region in a). FINE thus properly segments the organs, as show in d). }
    \label{fig:intro}
\end{figure}

Recent attempts have been made to apply transformers for 3D medical image segmentation. nnFormer \cite{zhou2021nnformer} is a 3D extension of SWIN~\cite{liu2021Swin} with a U-shape architecture.
~One limitation
~relates to the inherent compromise in multi-resolution, which prevents it from jointly using global context and high-resolution information. In \cite{zhou2021nnformer}, only local context is leveraged in the highest-resolution features maps. Models using deformable transformers such as CoTr \cite{cotr} are able to leverage sparse global-context.
A strong limitation shared by nnFormer and CoTr is that they cannot process large volumes at once and must rely on training the segmentation model on local 3D random crops. Consequently
, full global contextual information is unavailable and positional encoding can be meaningless. On BCV \cite{synapse}, cropped patch size is about $128\times 128 \times 64$ which only covers about $6\%$ of the original volume.

This paper introduces the Full resolutIoN mEmory (FINE) transformer. This is, to the best of our knowledge, the first attempt at processing full-range interactions at all resolution levels with transformers for 3D medical image segmentation. 
To achieve this goal, memory tokens are used to indirectly enable full-range interactions between all volume elements,
~even when training with 3D crops. Inside each 3D crop, FINE introduces memory tokens associated to local windows.
~A second level of localized memory is introduced at the volume level to enable full interactions between all 3D volume patches. We show that FINE outperforms state-of-the-art CNN, transformers, and hybrid methods on the 3D multi-organ BCV dataset \cite{synapse}.
\cref{fig:intro} illustrates the rationale of FINE to segment the red crossed kidney voxel in a). We can see that FINE's attention map covers the whole image, enabling to model long-range interactions between organs. In contrast, the receptive field of state-of-the art methods only cover a small portion of the volume, \eg the crop size (blue) for CoTr \cite{cotr} or the even smaller window's size (green) for nnFormer \cite{zhou2021nnformer} at the highest resolution level.

\section{FINE transfomer}

\begin{figure}
\centering
\includegraphics[width=0.8\linewidth]{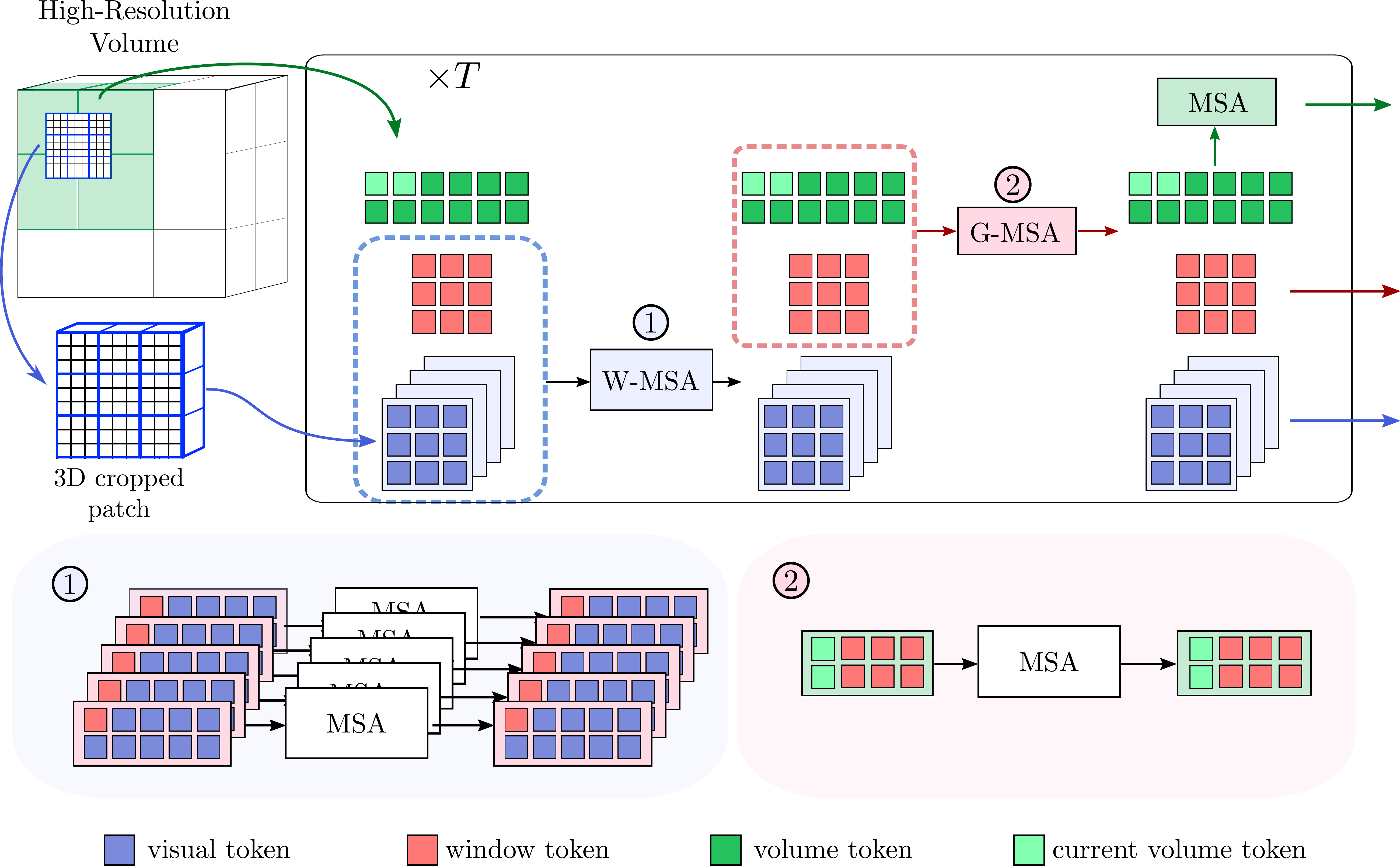}
\caption{To segment the cropped patch in blue and model global context, two level of memory tokens are introduced: window (red) and volume (green) tokens. First, the blue crop is divided into windows over which Multi-head Self-Attention (MSA) is performed in parallel. For each window, the sequence of visual tokens (blue) is augmented with a specific window token. Second, the local information embedded into each window token is shared between all window tokens and volume tokens intersecting with the crop (light green). Finally, high-level information is shared between all volume tokens).  } 
\label{fig:globals}
\end{figure}

In this section, we detail the FINE transformer for 3D segmentation of medical images leveraging global context and full resolution information, as shown in \cref{fig:globals}.
~FINE is generic and can be added to most multi-resolution transformer backbones~\cite{liu2021Swin,cao2021swinunet,zhou2021nnformer}. We chose to incorporate it to nnFormer \cite{zhou2021nnformer}, a strong model for 3D segmentation (see supplementary material).

\subsection{Memory tokens for high resolution semantic segmentation}

The core idea in FINE is to introduce memory tokens to enable full-range interactions between all voxels at all resolution levels with random cropping. We introduce memory tokens at two levels.

\textbf{Window tokens.} Multiple memory tokens can represent embeddings specific to regions of the feature maps \cite{hwang2021video}. Sharing these representations can thus leverage the small receptive field associated with window transformers' early stages. In this optic, we add specific memory tokens to the sequence of visual tokens associated with each window. We chose to call them window tokens to avoid any confusion.

\textbf{Volume tokens.} When dealing with high-resolution volumes, random cropping is a common training strategy. This approach is a source of limitation as only a portion of the spatial context is known by the model. Worse, no efficient positional embedding can be injected as the model has no complete knowledge of the body structure. Our memory tokens overcome this issue by keeping track of the observed part of the volume. These volume tokens are associated with each element of a grid covering the entire volume and called by the transformer blocs when performing the segmentation of a cropped patch. As can be seen \cref{fig:att_visu}, the volume tokens induce a positional encoding learned over the entire volume. 

\textbf{Discussion on memory tokens.}
The window and volume tokens can be seen as a generalisation of the class tokens used in NLP or image classification \cite{devlin2018bert,dosovitskiy2020vit,zhang2021vil}. In image classification, one class token is used as a global representation of the input and sent to the classifier. In semantic segmentation, more local information needs to be preserved which requires more memory tokens. 

\subsection{Memory based global context}
Each level of memory token in FINE is related to a subdivision of the input. These memory tokens and their corresponding regions are illustrated in \cref{fig:globals}. The high-resolution volume is divided into $M$ sub-volumes. Each sub-volume is associated to a sequence of $N_w$ $c$-dimensional volume tokens $\V w \in \mathbb{R}^{(M \cdot N_w) \times c}$. A 3D patch $\V p$ in input of the model is divided into $N$ windows. Each window is  associated to a sequence of $N_v$ window tokens $\V v \in \mathbb{R}^{(N \cdot N_v) \times c}$. A window is composed of $N_u$ visual tokens $\V u \in \mathbb{R}^{(N\cdot N_u) \times c}$ which are the finest subdivision level. 

In \cref{fig:globals}, volume, window and visual tokens are indicated in green, red, and blue respectively. First, Multi-head Self-Attention (MSA) is performed for each window over the merged sequence of visual and window tokens. Given a sequence of visual tokens \textit{ie.} small patches, MSA is a combination of non-local mean for all the tokens in the sequence \cite{AIAYN}. This local operation is denoted as Window-MSA (W-MSA). Second, MSA is performed over the merged sequence of all window tokens and corresponding volume tokens to grasp long-range dependencies in the input patch. This operation is denoted as Global-MSA (G-MSA) and involve only the volume token corresponding to sub-volumes intersecting with $\V p$. Finally, full resolution attention is achieved by applying MSA over the sequence of volume tokens. Formally, the $t$-th FINE-transformer bloc is composed of the following three operations:

\begin{align}
    [\V u^t, \Vh v^t] &= \text{W-MSA}([\V u^{t-1}, \V{v}^{t-1}]), \notag \\
    [\V {v}^t, \V{\hat w}_\cap^t] &= \text{G-MSA}([\Vh v^{t-1}, \V{w}_\cap^{t-1}]),  \\
    \V{w}^t &= \text{MSA}(\V w^t). \notag
\end{align}
$\V w_\cap$ denotes the volume tokens corresponding to sub-volumes with a non null intersection with $\V p$ and $[\V x, \V y]$ stands for the concatenation of $\V x$ and $\V y$ along the first dimension.

\subsection{FINE Properties}

\indent\textbf{Full range interactions.} After two FINE transformer blocs, the memory tokens manage to capture global context in the entire volume. This global context can then be propagated to visual tokens from the current patch - see supplementary and \cref{fig:att_visu}.

\textbf{Complexity}
MSA has quadratic scaling with respect to the 3D patch dimensions while W-MSA complexity is linear with respect to the input size \cite{liu2021Swin}. FINE only adds a few memory tokens and its complexity is given by:  
\begin{align}
    \Omega(\text{FINE}(\V u,\V v, \V w)) =& \Omega(\text{W-MSA}(\V u, \V v)) + \Omega(\text{G-MSA}(\V v, \V w_\cap)) + \Omega(\text{MSA}(\V w)) \notag \\
                              =& 2c\bigl( N (N_u+2N_v)+N_{w_\cap} + MN_w \bigr)\bigl(2c+1\bigr).
\end{align}
$N_{w_\cap}$ is the number of sub-volumes intersecting with the input patch and can not exceed 8.
Only a small number of global tokens brings consistent improvements and we keep $N_v=N_w=1$. In these conditions, memory tokens are particularly efficient with a negligible complexity overhead compared to W-MSA. 

\section{Experiments}
The Synapse Multi-Atlas Labeling Beyond the Cranial Vault (BCV) \cite{synapse} dataset is used to compare performances.
This dataset comprises 30 CT abdominal images with 7 manually segmented organs per image as ground truth. The organs are spleen (Sp), kidneys (Ki), gallbladder (Gb), liver (Li), stomach (St), aorta (Ao) and pancreas (Pa). The baselines are classic convolutional methods in medical image segmentation \cite{ronneberger2015unet,oktay2018attention,milletari2016vnet,Isensee2020nnUNetAS} and recent state-of-the-art transformer networks \cite{cao2021swinunet,chen2021transunet,hatamizadeh2021unetr,cotr,zhou2021nnformer}.

\noindent
\subsection{Data preparation and FINE implementation} 
 All images are resampled to a same voxel spacing. The CT volumes in BCV are not centered, with strong variation along the $z$ (cranio-caudal)-axis. To deal with this issue, the memory tokens are constant along this direction. 
 The sub-volumes are thus reshaped with the same depth as the original volume.  FINE is implemented in Pytorch and trained using a single NVidia Tesla V100-32GB GPU. All training parameters (learning rate, number of epochs, data augmentations are provided in the supplementary material).
 Each training epoch has 250 iterations where a randomly cropped region of size $128\times 128 \times 64$ voxels is processed. 
The loss function combines multi-label Dice and cross-entropy losses, and it is optimized using SGD with a polynomial learning rate decay strategy. Deep-supervision is used during training, where the output at each decoder stage is used to predict a downsampled segmentation mask. To avoid random noise perturbation coming from unseen memory tokens during training (typically memory tokens from regions that have never been selected), a smooth warm-up of these tokens is used. This warm-up consists of masking unseen tokens such that they do not impact the attention or the gradient. 

\begin{table}[h!]
    \centering
    \begin{tabular}{l|cc|ccccccc}
        \hline
        \multirow{2}{*}{Method} & \multicolumn{2}{c|}{Average} & \multicolumn{7}{c}{Per organ dice score (\%)} \\

         {} & HD95 & DSC & Sp & Ki & Gb & Li & St & Ao & Pa \\

        \hline
        UNet  \cite{ronneberger2015unet}     & -    & 77.4 & 86.7 & 73.2 & 69.7 & 93.4 & 75.6 & 89.1 & 54.0 \\
        AttUNet \cite{oktay2018attention}   & -    & 78.3 & 87.3 & 74.6 & 68.9 & 93.6 & 75.8 & 89.6 & 58.0 \\
        VNet  \cite{milletari2016vnet}     & -    & 67.4 & 80.6 & 78.9 & 51.9 & 87.8 & 57.0 & 75.3 & 40.0 \\
        Swin-UNet \cite{cao2021swinunet} & 21.6 & 78.8 & 90.7 & 81.4 & 66.5 & 94.3 & 76.6 & 85.5 & 56.6 \\
        nnUNet  \cite{Isensee2020nnUNetAS}   & 10.5 & 87.0 & 91.9 & 86.9 & \textbf{71.8} & \textbf{97.2} & 85.3 & \textbf{93.0} & \textbf{83.0}\\
        TransUNet \cite{chen2021transunet}  & 31.7 & 84.3 & 88.8 & 84.9 & 72.0 & 95.5 & 84.2 & 90.7 & 74.0 \\
        UNETR  \cite{hatamizadeh2021unetr}    & 23.0 & 78.8 & 87.8 & 85.2 & 60.6 & 94.5 & 74.0 & 90.0 & 59.2 \\
        CoTr*  \cite{cotr}    & 11.1 & 85.7 & 93.4 & 86.7 & 66.8 & 96.6 & 83.0 & 92.6 & 80.6 \\
        nnFormer \cite{zhou2021nnformer}  & 9.9  & 86.6 & 90.5 & 86.4 & 70.2 & 96.8 & 86.8 & 92.0 & 83.3 \\
        FINE*      & \textbf{9.2}  & \textbf{87.1} & \textbf{95.5} & \textbf{87.4} & 66.5 & 97.0 & \textbf{89.5} & 91.3 & 82.5 \\
        
        \hline
    \end{tabular}
    \caption{Method comparison using the BCV dataset and the training / test split from \cite{zhou2021nnformer}. Average Dice scores are shown (DSC in \% - higher is better). The average and individual organ 95\% Hausdorff distances are also shown (HD95 in mm - lower is better). * denotes results trained by us using the authors' public code.}
    \label{tab:sota_bcv}
\end{table}

\subsection{Comparisons with state-of-the-art}
\textbf{Single fold comparison}
To fairly compare with reported SOTA results, the same single split of 18 training and 12 test images was used as detailed in \cite{zhou2021nnformer}. The results are provided in Table \ref{tab:sota_bcv}. FINE obtains the highest average Dice score of 87.1\%, which is superior to all other baselines. It also attains the best average 95\% Hausdorff distances (HD95) of 9.2mm. 
Note that the second best method in Dice (nnUNet) is largely below FINE in HD95 (10.5), and the the second best method in HD95 (nnFormer) has a large drop in Dice (86.6).

\begin{table}[h!]
    \centering
    \resizebox{\linewidth}{!}{
    \begin{tabular}{l|c||c|c|c|c|c|c|c}
        \hline
        Method & Average & Sp & Ki  & Gb  & Li & St & Ao & Pa\\
        \hline
        CoTr \cite{cotr} & $ 84.4 \pm 3.7 $  & $ 91.8 \pm 5.0 $ & $ 87.9 \pm 3.4 $ & $ 60.4 \pm 10.0 $ & $ 95.7 \pm 1.4 $ & $ 84.8 \pm 1.3 $ & $ \textbf{90.3} \pm 1.8 $ & $ 80.0 \pm 3.2 $ \\
        nnFormer \cite{zhou2021nnformer} & $ 84.6 \pm 3.6 $  & $ 90.5 \pm 6.1 $ & $ 87.9 \pm 3.3 $ & $ 63.3 \pm 8.1 $ & $ 95.7 \pm 1.7 $ & $ 86.4 \pm 0.8 $ & $ 89.1 \pm 2.0 $ & $ 79.5 \pm 3.5 $ \\
        FINE    & $ \textbf{86.3} \pm 3.0 $    & $ \textbf{94.4} \pm 1.9 $ & $ \textbf{90.5} \pm 4.3 $ & $ \textbf{65.9} \pm 7.8 $ & $ \textbf{96.0} \pm 1.1 $ & $ \textbf{87.9} \pm 1.2 $ & $ 89.4 \pm 1.7 $ & $ \textbf{80.2} \pm 2.8 $ \\
        \hline
        P-values & \multicolumn{4}{c}{FINE vs. Cotr : 3e-2} & \multicolumn{4}{c}{FINE vs. nnFormer : 5e-2} \\
        \hline
    \end{tabular}
    }
    \caption{Method comparison with SOTA transformer baselines (CoTr and nnFormer) using the BCV dataset and 5-fold cross validation. Results show mean and standard deviation of Dice (in \%) for each organ and the average Dice over all organs (higher is better). }
    \label{tab:bcv}
\end{table}

\noindent\textbf{5-fold cross-validation comparison}
5-fold cross-validation of 18 training and 12 test images was used to compare FINE with the public implementation of the leading transformer baselines (CoTr and nnFormer). The Dice score results are provided in Table \ref{tab:bcv}. FINE's average improvement is significant (more than 1.5 pt with the second baseline with low variance), and FINE gives the best results in 6 out of 7 organ segmentation. The statistical significance in Dice is measured with a paired 2-tailed t-test. The significance of FINE gains with respect to CoTr (3e-2) and nnFormer (5e-2) is confirmed.

\subsection{FINE Analysis}

\begin{table}[h!]
    \centering
    \begin{tabular}{l|cc|cc|ccccccc}
        \hline
               \multirow{2}{*}{Method} & \multirow{2}{*}{WT} & \multirow{2}{*}{VT} & \multicolumn{2}{c|}{Average} & \multicolumn{7}{c}{Per organ dice score} \\

         {} & {} & {} & HD95 & DSC & Sp & Ki & Gb & Li & St & Ao & Pa \\

        \hline
        nnFormer   \cite{zhou2021nnformer}           &  0 &  0 & 8.0 & 86.2 & 96.0 & 94.2 & 57.2 & 96.5 & 87.2 & 89.5 & 82.5 \\
        \hline
        \multirow{2}{*}{FINE} &  $\checkmark$ &  0 & 7.7 & 86.6 & 95.7 & 94.2 & 60.9 & \textbf{96.8} & 85.1 & 90.0 & \textbf{83.8} \\
        {}                    &  $\checkmark$ &  $\checkmark$ & \textbf{5.2 } & \textbf{87.1} & \textbf{96.2} & \textbf{94.5} & \textbf{61.5} & \textbf{96.8} & \textbf{87.3} & \textbf{90.3} & 83.0\\
        \hline
    \end{tabular}
    \caption{Ablation study of the impact of different tokens on BCV dataset. The metrics are Dice score (DSC in \%) for all organs and in average, and the 95\% Hausdorff distance (HD95 in mm). WT: Window tokens. VT: Volume tokens. }
    \label{tab:abla_tok}
\end{table}

\begin{wraptable}{l}{0.38\textwidth}
    \centering
    \begin{tabular}{l|cc}
        \hline
        Method &Memory (GB) \\
        \hline
        nnUNet \cite{Isensee2020nnUNetAS}  &  8.6 \\
        CoTr  \cite{cotr}    & 7.62 \\
        nnFormer \cite{zhou2021nnformer} &  7.73 \\
        FINE     &  8.05 \\
        \hline
    \end{tabular}
    \caption{Models memory consumption during training.  }
    \label{tab:mod_technical}
\end{wraptable}

\noindent\textbf{Ablation study}
To show the impact of the different tokens in FINE, an ablation study is presented in Table \ref{tab:abla_tok}. Three variations of FINE are compared: FINE without tokens, which is equivalent to the nnFormer method; FINE with window tokens but without volume tokens, and FINE with window and volume tokens (default). The results shows that the window tokens generally help to better segment small and difficult organs like the pancreas (Pa) and gallbladder (Gb). The use of window tokens leads to an increase in average Dice by +0.4 points. Furthermore, adding volume tokens increases performance further (average Dice increase of +0.5 points, and average HD95 reduction from 7.7mm to 5.2mm).


\noindent\textbf{FINE complexity} The 
memory consumption of FINE compared to baselines is shown in Table \ref{tab:mod_technical}. FINE has very low overhead compared to CoTr and nnFormer. In addition, FINE has even a lower consumption than nnUNet.

\begin{figure}[h!]
    \centering
    \includegraphics[width=0.7\linewidth]{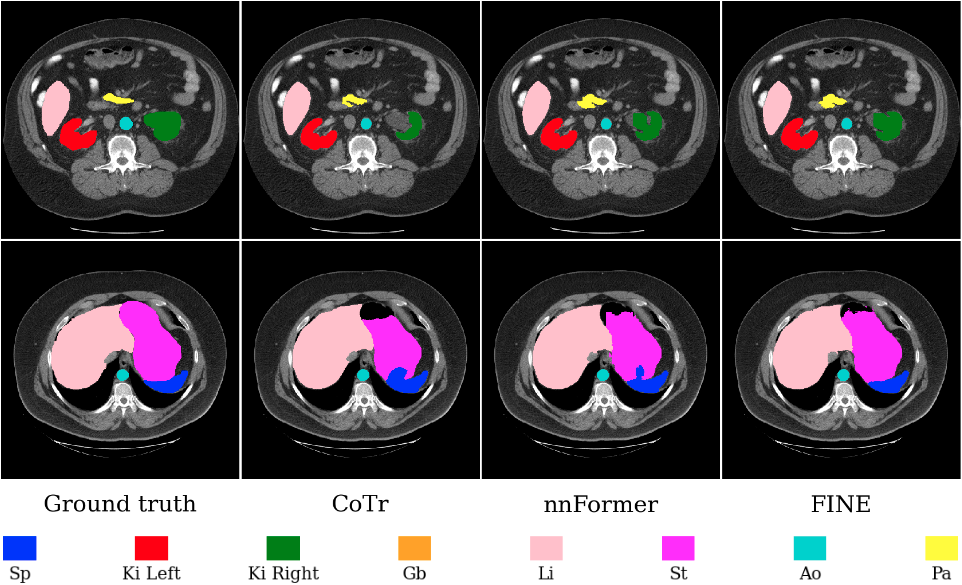}
    \caption{Visualisation of organs segmentation by FINE compared to state-of-the-art methods on BCV. We can qualitatively see how the full context and high-resolution in FINE help in performing accurate segmentation.}
    \label{fig:seg_visu}
\end{figure}

\begin{figure}[h!]
    \centering
    \includegraphics[width=0.7\linewidth]{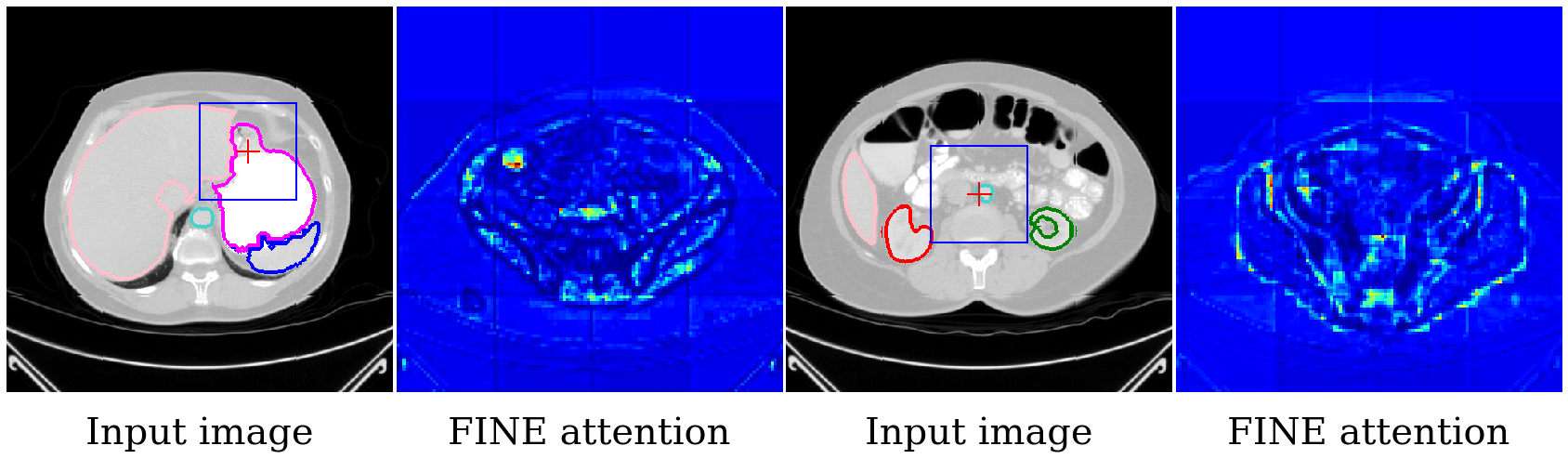}
    \caption{Visualisation of attention maps of FINE on a BCV segmentation example. The blue rectangle is the sub-volume for which the attention has been calculated.}
    \label{fig:att_visu}
\end{figure}

\noindent\textbf{Visualizations}
Visualizations of segmentation results from FINE compared to CoTr and nnFormer are presented in Figure \ref{fig:seg_visu}. All models produce compelling results compared to the ground truth, but one can clearly see differences and especially an improved segmentation from FINE of the spleen. Visualizations of FINE attention maps are provided in figure \ref{fig:att_visu}. These attention maps show that FINE is able to leverage context information from the complete image. The left example shows that there is attention with organs and tissues outside of the crop region (blue rectangle). Furthermore, the right example shows attention from different borders and bones like the spine, which give a strong positional information to the model.

\section{Conclusion}
We have presented FINE: the first transformer architecture that allows all available contextual information to be used for automatic segmentation of high-resolution 3D medical images. The technique, using two levels of memory tokens (window and volume), is applicable for any transformer architecture. Results show that FINE improves over recent and state-of-the-art transformers models. Our future work will involve the study of FINE in other modalities such as MRI or US images, as well as for other medical image tasks.

\newpage
\bibliographystyle{splncs04}
\bibliography{samplepaper}

\end{document}